\documentclass[12pt,twoside]{article}   
\usepackage[super,sort,comma]{natbib}

\usepackage{fancyhdr}		

\usepackage[section]{placeins}   %

\usepackage{graphicx}

\makeatletter \renewcommand\@biblabel[1]{$^{#1}$} \makeatother
\newcommand{\cen}[1]{\begin{center} #1 \end{center}}

\usepackage[mathlines]{lineno}

\usepackage{hyperref}
\hypersetup{ colorlinks,
    citecolor=blue,
    filecolor=blue,
    linkcolor=blue,
    urlcolor=blue
}

\usepackage{xcolor}

\definecolor{gray}{rgb}{0.6,0.6,0.6}
\definecolor{red}{rgb}{0.85,0,0}
\definecolor{green}{rgb}{0,0.85,0}
\definecolor{blue}{rgb}{0,0,0.85}
\definecolor{beige}{rgb}{0.92,0.87,0.78}
\usepackage[all]{hypcap}    

\usepackage{amsmath}
\usepackage{stfloats}
\usepackage{color}

\usepackage{algorithm}  
\usepackage{algorithmicx}  
\usepackage[noend]{algpseudocode}

\usepackage{easyReview}        
\newcommand{\LMM}{BLRM}
\usepackage{enumerate}
\usepackage{multirow}
\usepackage{lineno}

\let\oldequation\equation
\let\oldendequation\endequation

\renewenvironment{equation}{\linenomathNonumbers\oldequation}{\oldendequation\endlinenomath}
     
\begin{document}

\cen{\sf {\Large {\bfseries Bayesian Statistics Guided Label Refurbishment Mechanism: Mitigating Label Noise in Medical Image Classification } \\  
\vspace*{10mm}
Mengdi Gao\textsuperscript{1,2,3,4}, Ximeng Feng\textsuperscript{1,2,3,4}, Mufeng Geng\textsuperscript{1,2,3,4}, Zhe Jiang\textsuperscript{1,2,3,4}, Lei Zhu\textsuperscript{1,2,3,4}, Xiangxi Meng\textsuperscript{5}, Chuanqing Zhou\textsuperscript{4}, Qiushi Ren\textsuperscript{1,2,3,4}, Yanye Lu\textsuperscript{2,3}} \\
\textsuperscript{1} Department of Biomedical Engineering, College of Future Technology, Peking University, Beijing 100871, China \\
\textsuperscript{2} Institute of Medical Technology, Peking University Health Science Center, Peking University, Beijing 100191, China\\
\textsuperscript{3} Institute of Biomedical Engineering, Peking University Shenzhen Graduate School, Shenzhen 518055, China\\
\textsuperscript{4} Institute of Biomedical Engineering, Shenzhen Bay Laboratory 5F, Shenzhen 518071, China\\
\textsuperscript{5} Key Laboratory of Carcinogenesis and Translational Research (Ministry of Education), Key Laboratory for Research and Evaluation of Radiopharmaceuticals (National Medical Products Administration), Department of Nuclear Medicine, Beijing Cancer Hospital \& Institute, Beijing, China 
\vspace{5mm}\\
 Running Title: Bayesian-Guided Label Refurbishment\\
}

\pagenumbering{roman}
\setcounter{page}{1}
\pagestyle{plain}
Corresponding author: Yanye Lu, e-mail: yanye.lu@pku.edu.cn \\

\begin{abstract}
\noindent {\bf Purpose:} Deep neural networks (DNNs) have been widely applied in medical image classification, benefiting from its powerful mapping capability among medical images. However, these existing deep learning-based methods depend on an enormous amount of carefully labeled images. Meanwhile, noise is inevitably introduced in the labeling process, degrading the performance of models. Hence, it is significant to devise robust training strategies to mitigate label noise in the medical image classification tasks.  \\
{\bf Methods:} In this work, we propose a novel Bayesian statistics guided label refurbishment mechanism (\LMM) for DNNs to prevent overfitting noisy images. \LMM~utilizes maximum a posteriori probability (MAP) in the Bayesian statistics and {the} exponentially time-weighted technique to selectively correct the labels of noisy images. The training images are purified gradually with the training epochs when \LMM~is activated, further improving classification performance.\\
{\bf Results:} Comprehensive experiments on both synthetic {noisy} images (public OCT \& Messidor datasets) and real-world {noisy} images (ANIMAL-10N) demonstrate that \LMM~refurbishes the noisy labels selectively, curbing the adverse effects of noisy data. Also, the anti-noise \LMM~integrated with DNNs are effective at different noise ratio and are independent of backbone DNN architectures. In addition, \LMM~is superior to state-of-the-art comparative methods of anti-noise. \\
{\bf Conclusions:} These investigations indicate that the proposed \LMM~is well capable of mitigating label noise in medical image classification tasks. \\

\end{abstract}

\textbf{KEYWORDS}: deep learning, noisy label, label refurbishment, medical image classification

\newpage     

\tableofcontents

\newpage

\setlength{\baselineskip}{0.7cm}      

\pagenumbering{arabic}
\setcounter{page}{1}
\pagestyle{fancy}
\section{Introduction}

Over the past decade, learning-based computer vision algorithms have been widely explored and have contributed to medical imaging research. Artificial intelligence especially DNNs have exhibited impressive performance in numerous medical imaging tasks, such as image classification \cite{gulshan2016development}, object detection \cite{yan2018deeplesion}, semantic segmentation \cite{fan2020inf} and image synthesis \cite{shin2018medical}, etc. While the remarkable medical image classification results achieved by various deep learning methods \cite{xiao2015learning,gulshan2016development,kermany2018identifying}, highly depend on large-scale datasets with reliable labels. However, it is costly and time-consuming to collect and label datasets. In the medical scenarios, the annotations of images strongly rely on professional and experienced specialists and double-blinded even muti-blinded annotations are necessary. 
Meanwhile, some ineluctable labeling mistakes from annotators may generate noisy labels that may {deviate} from ground-truth labels. It has been reported that the ratio of noisy labels in {real-world datasets} ranges from 8.0\% to 38.5\% \cite{xiao2015learning,li2017webvision,lee2018cleannet,song2019selfie,song2020learning}. 


Fig. 1 describes the performence of optical coherence tomography images classification on noisy labeled training datasets. The details of dataset can be referred in Dataset description of section D. 
The noise ratio ranges from 0 to 40\% with step 10\%. The train accuracies,  converging to optimal value 100\% or nearly 100\% in Fig. 1 (a), prove that DNNs  can easily fit the entire training dataset with any ratio of corrupted labels. This viewpoint was also demonstrated in the existing computer vision literatures  \cite{park2020provable,jiang2018mentornet}. Fig. 1 (c) and (d) 
indicate that DNNs are capable of memorizing noise data, leading to poor generalization on the test dataset and the generalization performance  degrades as the noise ratio increases. Zhang et al. \cite{zhang2016understanding} has also drawn the similar conclusions in the classic image recognition tasks. In addition, the curves of test accuracy, especially at the heavy noise rate (30\% and 40\%), shows initially increased and then an downward trend which supports that DNNs tend to prioritize learning simple patterns first then memorize the remaining data including noisy data \cite{park2020provable,jiang2018mentornet}. Hence, it is meaningful to mitigate label noise to enhance the generalization performance of DNNs in medical image classification.  
Hence, the critical issue is how to train DNNs robustly even {with} noisy labels in the training data. 
Unfortunately, the above issue has not beeen settled completely through the popular {regularization} techniques, {such as} data augmentation\cite{shorten2019survey}, dropout\cite{srivastava2014dropout}, weight decay\cite{krogh1992simple}, and batch normalization\cite{ioffe2015batch}. 

As summarized by Song et al. \cite{song2020learning}, the noisy label problem has been addressed based on deep learning in five ways. 1) Sample selection: sample selection \cite{jiang2018mentornet,yu2019does,nguyen2019self} aimed to identify true-label samples from noisy training data. MentorNet \cite{jiang2018mentornet} introduced a collaborative learning paradigm where a pre-trained MentorNet would supervise the training of StudentNet. MentorNet guided StudentNet to focus on the sample where the label was probably correct, based on the low-loss trick. Sample selection excludes unreliable samples according to designed selection criterion, but it eliminates {obscure} yet useful training samples as well.
2) Robust architecture: robust architecture \cite{xiao2015learning,goldberger2016training,yao2018deep} attempted to design a new dedicated architecture or added a noise adaptation layer at the top of the SoftMax layer. However, the dedicated architecture lacks flexibility for extending to other architectures and {the} noise adaptation layer hinders a model’s generalization to complex label noise.
3) Robust regularization: robust regularization \cite{shorten2019survey,srivastava2014dropout,krogh1992simple,ioffe2015batch} aimed to enforce a DNN to overfit less to false-labeled samples. This technique introduces additional hyperparameters {and is} sensitive to {both} noise and data type{,} but unable to promote the model performance remarkably.
4) Robust loss function: robust loss function \cite{ghosh2017robust,zhang2018generalized,wang2019symmetric,lyu2019curriculum} aimed to modify the loss function and achieved a small risk for unseen clean data with the presence of noisy labels in the training data. Symmetric cross entropy (SCE) \cite{wang2019symmetric} was to integrate a noise tolerance term, namely reverse cross entropy loss, into the standard categorical cross entropy (CCE) loss. Nevertheless, the defect of the kind of technique is that it cannot combat the heavy and complex noise. 
5) Loss adjustment: loss adjustment \cite{patrini2017making,hendrycks2018using,wang2017multiclass,zhang2020learn} was intended to {mitigate} the adverse effects of noisy labels by adjusting the loss of training samples before updating the DNNs. Active bias \cite{chang2017active} emphasized hard samples with inconsistent label predictions and took prediction
variances as the weights during training.
Despite a full exploration of the training data when adjusting the loss of each sample, the error incurred by false correction is accumulated, particularly when the number{s} of {classes} and noise samples {are} large. In addition, SELectively reFurbIsh unclEan samples (SELFIE)  is a hybrid approach that combined advantages of both sample selection and loss adjustment \cite{song2019selfie}. The refurbished label of a training sample is determined by the most frequently predicted label for previous certain epochs when the sample satisfies the refurbished condition in SELFIE. Although it reduces the possibility of false correction while exploiting the full training data, it deals with predicted labels during time-period equally.


In this work, we propose a novel anti-noise training method (\LMM) which can integrate with DNNs, aiming to selectively refurbish noisy labels of the training data and update model parameters with the refurbished data and clean data. This \LMM~combines the advantages of both loss adjustment and sample selection.
\LMM~determines the refurbished label of each sample through Bayesian statistics on predicted labels during the latest $T$ epochs. 
We hypothesize that the predicted labels, derived from the well-trained models at the epoch of stable rising in performance before overfitting situation, are more credible. Based on the hypothesis, we propose an exponentially time-weighted technique to the predicted labels during the latest $T$ epochs, referring to from $(k-T)^{th}$ to $(k-1)^{th}$ epochs when refurbished label of a sample at the $k^{th}$ epoch is counted. Then, the refurbished label is calculated through counting exponentially time-weighted predicted labels from $(k-T)^{th}$ to $(k-1)^{th}$ epochs and the predicted label at the $k^{th}$ epoch according to the maximum a posteriori probability (MAP) in the Bayesian statistics (as illustrated in Fig. 2).
In addition, the start-up condition of \LMM~is put forward to prevent refurbishing labels prematurely, which hinges on the gradient of accuracy and loss of test dataset. 
We validated the superiority of \LMM~on public OCT \cite{kermany2018identifying} and Messidor \cite{messidor2014messidor} dataset, enhancing the medical image classification under simulated and different noisy rate. 
Besides, \LMM~also contributed to the performance improvement of classification on a real-world and natural images set ANIMAL-10N \cite{song2019selfie}.
In summary, the contributions of this paper are as follows: 
\begin{enumerate}
	\item Our propoed anti-noise \LMM~integrated with DNNs is effective to mitigate the label noise. \LMM~combines exponentially time-weighted and MAP in the Bayesian statistics techniques to purify the actual training data. Also, the start-up condition of \LMM~ is analyzed.
	\item \LMM~is proved to be independent of the backbone DNN and resistant to different noise rate from 10\% to 40\%.   
	\item Not only simulated noisy public available OCT dataset \cite{kermany2018identifying} and Messidor dataset \cite{messidor2014messidor}, but also a real-world ANIMAL-10N dataset validate the superiority of \LMM.
	\item  Both binary classification and muliti-class classification of medical images are conducted to demonstrate the efficiency of \LMM~in curbing adverse effects of noisy labels. 
\end{enumerate}

The rest of the paper is organized as follows. Section II presents the background and details of the proposed approach. Section III shows the results of {the proposed methods on} three public datasets. Section IV discusses relevant issues. Finally, Section V concludes the paper.

\section{MATERIAL AND METHODS}
\subsection{Problem Definition and Algorithm Description}
In a typical medical image classification task, the training dataset $D = \{(x_{i}, y_{i}) \vert 1 \le i \le N\}$ consisting of the sample $x_{i}$, and corresponding label $y_{i}$ is collected. {The paired} $(x_{i}, y_{i})$ is i.d.d. The goal of the task is to learn a function which maps the feature space of $x_{i}$ to the ground-truth label space $y_{i}$.
In this work, a mapping function $y = F(x; \theta)$ is learned by {DNN} to classify the retinal fundus images, where $\theta$ is the parameter of $F$. The parameter $\theta$ is learned by minimizing the empirical risk loss function and is updated along the descent direction of the expected loss on the mini-batch samples $B$, where $B$ is the subset of $D$. 

\begin{algorithm}[t!]    
	\caption{Pseudocode of the proposed \LMM} 
	INPUT: $D$: train data, $B$: mini-batch data, $\epsilon$: uncertainty threshold, $\gamma$: noise rate, \\ $T$: window width
	
	OUTPUT: $\theta_{t}$: model parameters, $\psi$: refurbished data, $C$: clean data
	\begin{algorithmic}[1] 
		\State $\psi$ $\gets$ $\emptyset$
		\State $t$ $\gets$ $1$
		\State $\theta_{t}$ $\gets$ Initialize the model parameters;    
		\For{$i=1$ to $epochs$}
		\For{$j=1$ to $\vert D \vert / \vert B \vert $}
		\State Extract a mini-batch $B$ from $D$;
		\If{$i$ belongs to warm-up period}  		
		\State $\theta_{t+1}=\theta_{t}-\alpha \bigtriangledown \frac{1}{\vert B \vert}\sum_{x\in B}L(x,y;\theta_{t})$  \qquad /* $\theta_{t}$ updated by Eq. (1)*/
		\Else{ $i$ reaches start-up condition of \LMM}
		\State $C$ $\gets$ $(1-\gamma)*100 \%$ of low-loss data in $B$ \qquad /* Clean samples selection*/	
		\For{each $x$ $\in$ $B$}
		\If{$Entropy$ ($x$,$T$) $\le$ $\epsilon$ or $x$ $\in$ $\psi$}
		\State Calculate $y^{\mathrm{refurb}}$ based on \LMM~ 
		\State $\psi$ $\gets$ $\psi$  $\cup$ ($x$,$y^{\mathrm{refurb}}$)
		\State /* $\theta_{t}$ updated by Eq. (2)*/ 
		\State $\theta_{t+1} = \theta_{t}-\alpha \bigtriangledown \frac{\sum_{x\in \psi}L(x,y^{\mathrm{refurb}};\theta_{t}) +\sum_{x\in C \cap \psi^{\prime}}L(x,y;\theta_{t}) }{\vert \psi \cup C \vert} $ 		
		\State $t$ $\gets$ $t$ + $1$
		\EndIf
		\EndFor
		\EndIf
		\EndFor 
		\EndFor  
		\State return $\theta_{t}$, $\psi$
	\end{algorithmic}  
\end{algorithm} 

\begin{equation}
	\theta_{t+1}=\theta_{t}-\alpha \bigtriangledown \frac{1}{\vert B \vert}\sum_{x\in B}L(x,y;\theta_{t}),  
	\label{eq_1}
\end{equation}
where $\alpha$ and $L$ are the learning rate and loss function respectively. Considering the possible corruption of sample labels in many real-world scenarios, this study aims to modify the update equation (\ref{eq_1}) to render the network more robust{ness} on noisy labels. 
Algorithm 1 describes the overall procedure of our proposed \LMM~to handle the noisy labels. First, in the warm-up period, {which is} the initial $n$ epochs of training, the network is trained on the whole training dataset in the default manner as shown in equation (\ref{eq_1}) (Algorithm 1, Lines 6–8). Notwithstanding the existence of noisy data, the memorization effects \cite{park2020provable,jiang2018mentornet} indicate that DNNs will initially `memorize' the training samples with clean labels and then those of noisy labels. Subsequently, the start-up condition of label refurbishment mechanism is reached, {and} the training samples in the mini-batch $B$ are separated into clean samples, refurbished samples and the remaining samples.  
Let $C \subset B$ be the clean samples and $\psi \subset B$ be the refurbished samples. Subset $C$ covers $(1 - \gamma) \times 100\%$ {of} low-loss instances\cite{jiang2018mentornet} and $\gamma$ is the noise rate (Algorithm 1, Lines 9–10). If $\gamma$ is unknown, it can be reconstructed through cross-validation \cite{liu2015classification}. 
{In} this period, each train sample is identified through checking the predictive uncertainty that uses the entropy to measure the consistency of label prediction in the $T$ epochs (Algorithm 1, Line 12). The detailed calculation of entropy can refer to the previous research \cite{song2019selfie}.
Our proposed \LMM~is applied to determine the refurbished labels of samples in $\psi$ (Algorithm 1, Line 13). Then the refurbished samples are aggregated into $\psi$ for reuse (Algorithm 1, Line 14).
{Notably}, the intersection of $C$ and $\psi$ is not necessarily {a} nonempty set. If a sample $x\in \psi \cap C$, being refurbished precedes being clean because mislabeled instances could be included even in $C$. The parameters $\theta$ of the DNN will be updated based on the clean samples along with refurbished samples. We correct the backward loss of the refurbished sample $x\in \psi$ by replacing its corrupted label $y$ with the refurbished {label} $y^{\mathrm{refurb}}$ and backpropagate the losses for the refurbished and clean samples to update the network (Algorithm 1, Lines 15-16), which can be described as:
\begin{equation}
	\begin{split}
		\theta_{t+1}=\theta_{t}-\alpha \bigtriangledown( \frac{1}{\vert \psi \cup C \vert} ( \sum_{x\in \psi}L(x,y^{\mathrm{refurb}};\theta_{t}) +\sum_{x\in C \cap \psi^{\prime}}L(x,y;\theta_{t}) )),
	\end{split}
	\label{eq_2}
\end{equation}
where $\psi^{\prime}$ represents the complement set of $\psi$.

\subsection{Techniques of Label Refurbishment Mechanism}
Fig. 2 describes the overall procedure of our proposed Bayesian statistics guided label refurbishment mechanism. Each mini-batch of medical images are fed into {the} base DNN to train {the} image classification model. Through the training process, predicted labels with corresponding probabilities of {all} training samples at each epoch are recorded to calculate likelihood function for the optimal label {selection}. At the early stage of training, original given labels should be used for loss calculation and start-up condition of \LMM~should be judged simultaneously. When the start-up condition of \LMM~is reached,{the} optimal label selection module will be activated and calculates the estimated labels $y^{\mathrm{refurb}}$ to replace the original given labels $y$. The technical details are illustrated {in} the following parts.

\textbf{Start-up condition of label refurbishment mechanism.} Before \LMM~takes into effect, a warm-up period is necessary for ensuring the performance of the training model reaching into relatively steady state. In the warm-up stage of training, the performance of the model is unstable or under-fitting, which leads to large deviations of the computation results of refurbished labels. Therefore, an appropriate start-up condition of \LMM~should be devised carefully, which can prevent the refurbished labels from fluctuating or unchanging unacceptably. In this study, we propose two prerequisites to activate \LMM. 
First, the average loss value of samples in validation dataset should step into a range $[L_{a}, L_{b}]$, in which our training model may acquire the best performance on validation dataset. Obviously, the value of $L_{a}$ is zero in an ideal situation. As the output of the last layer should be normalized by {a} SoftMax function, the {minimum} probability of the {g}round {t}ruth (GT) label for a {correct predicted sample} {would be} no less than $1/M$, where $M$ is the number of categories. According to the formula of cross-entropy, $L_{b}$ can be calculated as :
\begin{equation}
	\begin{split}
		L_{b}=\max \limits_{(1/M<P_{\mathrm{GT}} \le 1)}{L\vert L=-ln(P_{\mathrm{GT}})}= -ln(1/M),
	\end{split}
	\label{eq_3}
\end{equation}
{where} $P_{\mathrm{GT}}$ represents the probability of the GT label. Second, our model has not been suspected of over-fitting or under-fitting. In the training process, the accuracy of validation dataset should satisfy following {condition},
\begin{equation}
	\rho_{\mathrm{val}} > 100\% - \gamma - \phi,
	\label{eq_4}
\end{equation}
where $\rho_{val}$ represents accuracy on validation dataset during training and $\gamma$ is the original noise rate of training dataset. $\phi$ is a hyperparameter that is named {as} the relaxation factor in our study and is used to control the base accuracy to trigger \LMM. In our study, the value of $\phi$ was pre-set to 5\%.

\textbf{Bayesian statistics for optimal label selection. }The Bayesian statistics formula is utilized in our study to estimate the optimal label for each sample, which is shown as follows:
\begin{equation}
	p(Y_{\theta^{(k)}} \vert Y_{\theta ^{(k-1)} \sim \theta^{(k-T)}})=\frac{p(Y_{\theta ^{(k-1)} \sim \theta^{(k-T)}} \vert Y_{\theta^{(k)}}).p(Y_{\theta^{(k)}})}{Z}.
	\label{eq_5}
\end{equation}
Inside: 
\begin{equation}
	Z=\sum_{Y_\theta ^(k)=1}^M p(Y_{\theta ^{(k-1)} \sim \theta^{(k-T)}} \vert Y_{\theta^{(k)}}).p(Y_{\theta^{(k)}}).
	\label{eq_6}
\end{equation}
\begin{equation}
	\sum_{i=1}^M p(Y_{\theta ^{(k)}} = i)=1,
	\label{eq_7}
\end{equation}
where $Y_{\theta^{(k)}} \in \{1,…,M\}$ expresses the label of a sample used in $k^{th}$ epoch, {and} $Y_{\theta ^{(k-1)} \sim \theta^{(k-T)}}$ represents the label sequences ranging from the $(k-T)^{th}$ and the $(k-1)^{th}$ epoch. Besides, $T$ and $Z$ stand for the window width epoch for Bayesian statistics and the normalized constant{,} respectively. 

The prior probability $p(Y_{\theta^{(k)}})$ is {obtained} by the training model of the $k^{th}$ epoch. We need to make sure what statistic{al parameter} to use to calculate the likelihood function $p(Y_{\theta ^{(k-1)} \sim \theta^{(k-T)}} \vert Y_{\theta^{(k)}})$. Referring to our hypothesis that the estimated label of a current sample is related to its past learning effects, we regard $p(Y_{\theta ^{(k-1)} \sim \theta^{(k-T)}} \vert Y_{\theta^{(k)}})$ as a weighted mean statistic to {incorporate} past knowledge into current label estimation. The likelihood function can be computed by {the} following formula:
\begin{equation}
	p(Y_{\theta ^{(k-1)} \sim \theta^{(k-T)}} \vert Y_{\theta^{(k)}})= \sum_{i=1}^T \omega_{i}.p(Y_{\theta^{(k+i-T-1)}}=Y_{\theta^{(k)}}).
	\label{eq_8}
\end{equation}
Inside:                                          
\begin{equation}
	\omega_{i}=\frac{e^{\frac{i}{\eta}}}{Z^{'}},
	\label{eq_9}
\end{equation}
where $\omega_{i}$ denotes the weight of one label of $i^{th}$ epoch in window $T${;} $Z^{'}$ is a normalization constant where $Z^{'} = \sum_{i=1}^T e^{\frac{i}{\eta}}${;} and $\eta$ is an adjustable parameter that controls the distribution of $\omega_{i}$. In this study, we assigned a bigger weight to a more recently learned label, {as} the latest knowledge has the greatest impact on the decisions. In our study, $\eta$ was equal to the length of window width.
Then the posterior probability $p(Y_{\theta ^{(k-1)} \sim \theta^{(k-T)}} \vert Y_{\theta^{(k)}})$ can be calculated. The refurbished {label} $y^{\mathrm{refurb}}$ is corresponding to the class index that maximized the posterior probability.

\subsection{Network Architecture and Implementation}
To validate the effectiveness of \LMM, we integrated \LMM~into the DNN architecture and compared the performance between the DNNs with and without \LMM~based on the same train and test dataset. With new network architectures constantly emerging, the {compatibility} of the proposed \LMM~with any type of DNNs is important. Flexibility ensures that the proposed method can quickly adapt to the different architecture{s}. In the comparative experiments, three {popular} DNNs (VGG-16 \cite{simonyan2014very}, Inception-V3 \cite{szegedy2016rethinking}, Resnet-50 \cite{he2016deep}) {were} used to demonstrate the flexibility of \LMM$\footnote{The code will be released at \href{https://github.com/neugmd/BLRM}{https://github.com/neugmd/BLRM}}$. The training procedure utilized the Adam optimizer with a learning rate of 0.0001, a cross-entropy loss function, and a minibatch size of 32. 


\subsection{Dataset Accumulation and Transformation}
\textbf{Dataset description.} In this study, we sought to develop an effective noise reduction technique (\LMM)~to enhance the performence of noisy medical image classification. The primary illustration of this technique involves optical coherence tomography (OCT) images of the retina. The public OCT dataset \cite{kermany2018identifying} covers choroidal neovascularization (short as 1), diabetic macular edema (2), drusen (3), and normal cases (0). We randomly sampled 1,000 images for each category as training dataset and utilized the available validation dataset (250 images for each category). 

The \LMM~ was also tested in DR retinal fundus images to validate the generalization of this technique across multiple imaging modalities. We performed the binary image classification tasks on a public dataset Messidor \cite{messidor2014messidor}. Detailed grading information is listed in Table 1 and NMA, NHE, and NNV refer to the number of microaneurysms, hemorrhages and neovascularization, respectively. Fundus images with DR0 and DR1 are categorized as routine referrals (short as 0). These conditions would demand regular follow-up. While fundus images with DR2 and DR3 are categorized as urgent referrals (short as 1) where the patients demand relatively urgent referral to ophthalmologists for timely treatment. We referred to the errata available and then deleted 13 duplicate images and adjusted labels of 4 images with inconsistent grading. 

\textbf{Noise injection.} As the both datasets contained only clean samples, we need to artificially corrupt sample labels to generate noisy labels. Frénay and Verleysen \cite{frenay2013classification} summarized the taxonomy of label noise in detail. As shown in Fig. 3 (a), noise transition matrix $T_{ij}$ describes the probability of ground-truth label $i$ being flipped to the noisy labels $j$. For $M$ classes, symmetry noise satisfies 
\begin{equation}
	T_{ij}=\frac{\gamma}{M-1},
	\label{eq_10}
\end{equation}
where a ground-truth label is flipped into other labels with equal probability and the noise rate $\gamma \in [0,1]$. 


In our work, symmetry noise was introduced {respectively} to construct researchable datasets with noise. {To evaluate the robustness on varying noise rates from light noise to heavy noise, according to the real-world noise rate,} we tested five noise rates, varying from 0 to 40\% with step 10\%{, to validate the robustness of our proposed method}.

\subsection{Quantitative Evaluation Metrics and Comparative Study}
\textbf{Quantitative evaluation metrics.} The performance of \LMM~is quantitatively evaluated by test accuracy. The test dataset has unbiased and clean samples that are not used in training. The test accuracy degrades drastically when the DNN overfited samples with noisy labels \cite{zhang2016understanding}. Furthermore, area under curve (AUC) is also calculated as the metric. Meanwhile, data purity could be utilized as an indicator of the proportion of samples with ground-truth labels in the whole train{ing} dataset.
\begin{equation}
	Data \; purity = \frac{\vert \{(x_{i},y_{i}) \in D: \tilde{y_{i}} = y_{i} \} \vert}{\vert D \vert},
	\label{eq_11}
\end{equation}
Where $D$ is the whole train{ing} dataset and $y_{i}$ is the GT label and $\tilde{y_{i}}$ is the resulting label of the {$i^{\mathrm{th}}$} samples in $D$. $y_{i}$ is either original label or refurbished label. Data purity may be updated after each epoch when \LMM~worked. Cohen's kappa (kappa) \cite{kvaalseth1989note} is further employed to measure the agreement between ground-truth labels and noisy labels of train{ing} dataset.

\textbf{Comparative study methods.} We compared our proposed method with a benchmark model (marked as Default) and four robust training algorithms (Coteaching \cite{han2018co}, JoCoR \cite{wei2020combating}, AdaCorr \cite{zheng2020error}, and SELFIE \cite{song2019selfie}). We re-implemented with the same network backbone architecture to ensure the fairness of the comparison. Although they were designed and evaluated for natural images classification tasks, we re-adapted them with fine-tune hyper-parameters on medical datasets for a fair comparision. Compared methods adapted different strategies to mitigate label noise in the medical image classification. Hence, we cannot compare the data purity of all the comparison methods. The compared methods include:
\begin{enumerate}
	\item \textbf{Default} This is the common training procedure without any processing strategy {of} the noisy labels.  
	\item \textbf{Coteaching} This is the method proposed by \cite{han2018co}. Coteaching selected the clean samples by the loss-based separation and adopted the co-training mechanism to confront noisy annotations. 
	\item \textbf{SELFIE} This is the method proposed by \cite{song2019selfie}. SELFIE combined loss correction with {the} sample selection strategy to improve the robustness.
	\item \textbf{JoCoR} This method \cite{wei2020combating} was to train two classifiers simultaneously with small-loss instances, using both regular supervised loss and co-regularized loss.  
	\item \textbf{AdaCorr} This anti-noise method \cite{zheng2020error} proposed a label correction algorithm to combat label noise. 
\end{enumerate}

\section{EXPERIMENT AND RESULTS}
We initially verified the validity of the proposed \LMM~through four-class OCT images classification experiments based on public OCT dataset. After that, generalization was {proven} through binary DR images classification with the public {Messidor} dataset. Two classes refer to routine referral (DR0 and DR1) and urgent referral (DR2 and DR3) treatment group. The {sizes} of routine referral and urgent referral treatment group{s} are 550 and 380 for {the} train{ing} dataset, {and} 146 and 111 for test dataset, respectively. The division details of training and test set can be referred in Table 1.
All experiments were performed on an NVIDIA RTX3090 GPU with 24 GB of memory. In this work, we did not apply any data augmentation or pre-processing procedures.


\subsection{Hyperparameter Selection}
The proposed DNN with \LMM~receives two hyperparameters: the window width $T$ and the uncertainty threshold $\varepsilon$. To determine the optimal combination of hyperparameters, we trained Inception-V3 on the noisy OCT dataset at a rate of 40\% noise with two hyperparameters set in a grid {with} $T\in \{4,5,6\}$ and $\varepsilon \in \{0.05, 0.1, 0.15, 0.2, 0.25, 0.3, 0.35, 0.4, 0.45, 0.5\}$. Similarly, we repeated the grid-search experiments on the noisy Messidor dataset at a rate of 40\% noise with two hyperparameters set in a grid {with} $T\in \{5,10,15\}$ and $\varepsilon \in \{0.3, 0.325, 0.35, 0.375, 0.4, 0.425, 0.45\}$. Fig. 4 illustrates the test accuracy obtained by the grid search on the two noisy datasets, respectively. Regarding the uncertainty threshold, the best test accuracy cannot be achieved with both small and large thresholds. The performance generally involves a comprise between the correct{ly} refurbished samples in the $\psi$ and wrong{ly} refurbished samples in the $C$. The small threshold corresponds to the small rate of both the above two cases while the large threshold corresponds to the high rate of both the above two cases. In Fig. 4 (a), as for the window width, although there is no clear winner among the 4, 5, and 6, the $T$ of value 5 achieves the highest test accuracy when the threshold $\varepsilon$ was 0.25. Therefore, in the following experiments on the OCT dataset (Messidor dataset), we set the uncertainty threshold $\varepsilon$ to 0.25 (0.4) and the window width $T$ to 5 (5). 


\subsection{Performance of Four-class OCT Images Classification}
\textbf{The flexibility of the proposed \LMM.} The flexibility of \LMM~ensures the capability of supporting any type of DNN architectures. \LMM~{were separately integrated with} three popular DNNs (VGG-16, Resnet50, Inception-V3) to perform comparative experiments on the public OCT dataset. Here, the noise rate was fixed at 20\%. The detailed performance metrics are shown in Table 2. Three DNNs architecture integrated with \LMM~ all enhance  the generalization performance and reduce the influence of noisy labels. Compared with the benchmark VGG16, the test accuracy and data purity of VGG16-\LMM~improve from 0.82 and 79.90\% to 0.86 and 81.13\%, respectively. Resnet50 and Inception-V3 with \LMM~share the similar promoted trend. In general, the results prove that the proposed \LMM~can improve the performance of DNNs and the promotion is independent of specific models. 


\textbf{Tolerance to different proportions of noise.} We selected Inception-V3 as the benchmark model and set the noise {level} from 0 to 40\% with {a} step {size of} 10\% to test the tolerance of the proposed \LMM~to different proportions of noise. The comparison metrics of OCT dataset are listed in Table 3. We can see that the performance of the model without \LMM~becomes worser with the increase of noise ratio. 
In {the} case of no noise containing only clean training samples, \LMM~produces hardly any side effects to the performance of the well-trained models. Generally, under the noise level from 10\% to 40\%, \LMM~achieves better metrics than benchmark model on OCT dataset. For example, at the relatively heavy noise rate of 30\%, the data purity, ACC, and Kappa increase 2.80\%, 8.50\%, and 11.06\%, respectively. The Fig. 5 (a) and (b) illustrate the confusion matrices comparing test accuracy for OCT images classification without and with \LMM~at the noise of 40\%, respectively. We could observe that the \LMM~improves the test accuracy obviously and enhances performence of each category, especially for the normal cases (short as label 0).

\subsection{Performance of Two-class DR Images Classification}
\textbf{Generalization on the public Messidor dataset.} We also verified the generalization of \LMM~based on the corrupted Messidor dataset. Considering {that} the {size} of Messidor was relatively small (less than 1200), we did not {conduct} the image grading experiment, but trained the two-class DR images classification model based on the pretrained models derived from ImageNet. We still adopted Inception-V3 as the backbone network and utilized hyperparameters determined by grid-search experiments above. Messidor was corrupted with different noise {levels} of 10\%, 20\%, 30\% and 40\%, respectively. The test accuracy and train data purity of comparative experiments are displayed in Table 4. Similarly, the Inception-V3 with \LMM~weakenes the influences of noisy labels on the Messidor dataset. The Inception-V3 with \LMM~achieves improvement on test accuracy of 3.51\%, 6.62\%, 2.33\%, and 5.84\% under the noise rate raising from 10\% to 40\%, respectively. The Fig. 6 from (a) to (d) diaplsy the ROC curves using Inception-v3 with or without BLRM at each noise rate. The area under the ROC curve of Inception v3 with BLRM is larger at each noise rate , indicating BLRM improving the performance of Inception-v3. In addition, from the confusion matrices comparing test accuracy for DR images classification at the noise rate 40\% without and with \LMM~(Fig. 7), our proposed technique promotes the generalization performance of classification model. In brief, \LMM~optimizes the performance of models and improves purity of training data when learning from the noisy Messidor dataset.


\subsection{Comparison Study}	
The comparative methods including Default, Coteaching, SELFIE, JoCoR, and AdaCorr were employed on OCT dataset and Messidor dataset, respectively, to compare with our proposed \LMM. Fig. 8 shows the test accuracy of the compared methods with varying symmetry noise rates, ranging from 0 to 40\%. And the test accuracy of all methods at the noise rate of 30\% are summarized in Table 5. 

In the OCT dataset (Fig. 8 (a)), \LMM~surpasses reference methods at high noise rate, except JoCoR. We speculate that four-class learning task, identifying choroidal neovascularization, diabetic macular edema, drusen, and normal cases, possesses relatively sufficient training samples (1000 images for each category) relative to task complexity. Hence, JoCoR works well thanks to a joint loss with co-regularization for each training example. However, JoCoR has two classifiers to train which doubles the quantity of parameters, raising the computing resources and training time considerably. Both AdaCorr and our method belong to label refurbishment for the noisy data. Our proposed Bayesian statistics guided label refurbishment mechanism performs more stably while AdaCorr is inferior at the noise rate of 30\% and 40\%.  When compared with Default group, with the increase of noise ratio, the improvement of \LMM~is more significant and the maximum increment reaches 14\% when the noise rate is 40\%. Both Coteaching and SELFIE reduce the influence of noise label but there is no clear winner between Coteaching and SELFIE. SELFIE is superior to Coteaching in presence of light noise (10\% and 20\%) while SELFIE is inferior to Coteaching in presence of heavy noise (30\% and 40\%).
 
In the Messidor dataset (Fig. 8 (b)), as DNNs were trained based on the pretrained models derived from ImageNet due to the dearth of training data, the effectiveness of methods for resisting noisy label is relatively limited. Regarding Messidor dataset, our proposed method is superior to others generally, under high noise levels in particular. Although JoCoR outperforms ours at noise rate of 10\%, its performance degenerates sharply with the increase of noise ratio. Limited instances together with serious noisy labels interference in the Messidor dataset, caused JoCoR is indeed difficult to derive the joint loss including regular supervised part and co-regularized part, accurately.   
To sum up, our proposed \LMM~ can improve the robustness of models across different noise patterns and is superior to the comparison method in most cases.




\section{Discussion}
\subsection{Result with Realistic Noise}
In addition to manually induced noise, ANIMAL-10N \cite{song2019selfie} with realistic noise was further utilized to conduct image classification experiment{s} to validate the proposed \LMM~method. ANIMAL-10N consists of 10-class animal images with 50,000 training images and 5,000 test images. Notably, training dataset in ANIMAL-10N has realistic noise corrupted with noisy labels naturally by human mistakes and the noise rate is estimated at 8\%. And the test dataset in ANIMAL-10N is free from noisy labels. 

Further, the superiority of \LMM~is proved by comparing {with} five comparative methods (Default, Coteaching, SELFIE, JoCoR, and AdaCorr). As the correct ground-truth labels of {the} training dataset in ANIMAL-10N are unknown, the data purity and Kappa metrics of {the} training dataset cannot be calculated. The test accuracy is illustrated in the bar chart in Fig. 9. Our proposed \LMM~ranks first reaching 82.6\% and Default rank{s the} last reaching 79.4\%. \LMM~increases the accuracy by 1.6\% and 2.4\% compared with SELFIE and Coteaching, respectively. ANIMAL-10N possesses more categories and large-scale small-sized natural images. JoCoR is not competent for this scenario, with very limited promotion while AdaCorr achieves high test accuracy of 81.6\%, with merely 1\% lower than ours. In brief, \LMM~also works well when dealing with realistic noise. 


\subsection{\LMM~Applied in Weak Supervision Learning}
DNNs {for} image classification is sensitive to the quantity of training data to some extent {in the case of a fixed model}. When lacking training data, enriching training data is an effective way to improve the performance of the model. Self-training exploits unlabeled data with pseudo-label{s} to achieve better model performance. Inspired by self-training \cite{xie2020self}, we applies \LMM~to self-training to refurbish the pseudo-labels. We carried out three comparative experiments on the Messidor dataset and Fig. 10 illustrates the {test accuracy against the percentage of training dataset with ground-truth labels fed to DNNs}. The differences among {the} three groups of comparative experiments depend on the training data and strategies of the training model. Taking {$x$} value being 10\% for example, in {the} control group experiment, 10\% training data with {the} ground-truth label are utilized for training {the} DR image dichotomy task. While {the} self-training group contains the 10\% training dataset {above, as well as} 10\% dataset with pseudo-labels{, the s}elf-training with \LMM~group shares a similar training data pattern as {the} self-training and {applies} \LMM~to enhance {the} performance of the DNNs. In Fig. 10, all the three groups verify that test accuracy becomes higher along with the increase (from 10\% to 30\%) of the training dataset. {The s}elf-training group can indeed utilize the unlabeled samples and improve the test accuracy compared with the control group. Self-training with \LMM~{surpasses} the other two groups and increases 3.2\%, 5.9\%, 2.5\%, 2.9\%, and 2.8\% compared with the control group from the 10\% to 30\% percentage of {the} training dataset. It proves that \LMM~can leverage the features’ information of unlabeled data and boost performance in weak supervision learning task.


\subsection{The Interpretability of the Model}
The class activation maps (CAM) suggest that the sensitive areas which causes the high response of our model are consistent with the suspicious areas in a clinical diagnosis. CAM can prove the superiority of the DNNs with \LMM, which is shown as Fig. 11. The {leftmost} column shows the color retinal fundus images with corresponding ground-truth label in the upper left corner. The middle and {rightmost} column{s} are the CAMs with predicted labels and probabilities in the upper left corner from Inception-V3 with and without \LMM{,} respectively. In Fig. 11 (a), the predominant lesions including multiple hemorrhages lesions are located on the nasal side of the foveal location. Microaneurysms and hemorrhages lesions can be observed between superior and inferior vascular arcades in Fig. 11 (b). The model without \LMM~fails to detect the lesions, resulting in false-negative misjudgment {in both} cases. {However,} our proposed method can accurately localize the lesions and obtain correct positive predicted labels. Fig. 11 (c) is a negative case. The model without \LMM~treat the reflection of nerve fiber as lesions mistakenly, leading to {a} false-positive prediction with high probability of 0.9755. 

\subsection{Limitations and Future Direction}
This study still has some limitations. Firstly, our study initially carried out all the experiments based on the dataset with limited symmetry noise. In the {clinical settings}, the distribution of label noise is {unknown}. It{ is} worth exploring the taxonomy of label noise such as asymmetric (or label-dependent) noise\cite{frenay2013classification}. Asymmetric noise means that a ground-truth label is more likely to be mislabeled into a particular label, which is more reasonable in a real sense. Secondly, the features of noisy samples need to be further mined so that more noisy samples can be refurbished to correct ones and fewer clean samples can be refurbished to wrong ones. Thirdly, datasets from other medical modalities such as CT, MRI and PET can be utilized {to further} validate the effectiveness of our proposed \LMM. 

\section{Conclusion}
We propose a novel \LMM~for robust training {of} DNNs classification models with noisy labels. A selectively refurbished sample can be corrected through analyzing the former predicted labels with the Bayesian maximum {a} posteriori probability and exponentially time-weighted technology. We conducted extensive experiments of four-class OCT images classification and two-class DR images classification on public OCT and Messidor datasets with varying noise {levels}. Our experiment results show that \LMM~can improve the robustness of the DNNs when dealing with corrupted labels. \LMM~guides the network to avoid noise accumulation and allows it to take advantage of the full exploration of training data. In summary, the proposed \LMM~has demonstrated its capability {of} reducing the adverse effects of noisy labels in medical image classification based on deep learning.

\section*{Acknowledgement}
\addcontentsline{toc}{section}{\numberline{}Appendix}
This work was supported in part by Beijing Natural Science Foundation under Grant Z210008, in part by Shenzhen Science and Technology Program under Grant 1210318663, and in part by Shenzhen Nanshan Innovation and Business Development Grant.

\section*{Conflict of Interest}
\addcontentsline{toc}{section}{\numberline{}Appendix}
The authors have no conflicts to disclose.

\clearpage

\section*{References}
\addcontentsline{toc}{section}{\numberline{}References}
\vspace*{-20mm}





\bibliography{./example.bib}      

\begin{thebibliography}{10}

\bibitem{gulshan2016development}
V.~Gulshan et~al.,
\newblock Development and validation of a deep learning algorithm for detection
  of diabetic retinopathy in retinal fundus photographs,
\newblock Jama {\bf 316}, 2402--2410 (2016).

\bibitem{yan2018deeplesion}
K.~Yan, X.~Wang, L.~Lu, and R.~M. Summers,
\newblock DeepLesion: automated mining of large-scale lesion annotations and
  universal lesion detection with deep learning,
\newblock Journal of medical imaging {\bf 5}, 036501 (2018).

\bibitem{fan2020inf}
D.-P. Fan, T.~Zhou, G.-P. Ji, Y.~Zhou, G.~Chen, H.~Fu, J.~Shen, and L.~Shao,
\newblock Inf-net: Automatic covid-19 lung infection segmentation from ct
  images,
\newblock IEEE Transactions on Medical Imaging {\bf 39}, 2626--2637 (2020).

\bibitem{shin2018medical}
H.-C. Shin, N.~A. Tenenholtz, J.~K. Rogers, C.~G. Schwarz, M.~L. Senjem, J.~L.
  Gunter, K.~P. Andriole, and M.~Michalski,
\newblock Medical image synthesis for data augmentation and anonymization using
  generative adversarial networks,
\newblock in {\em International workshop on simulation and synthesis in medical
  imaging}, pages 1--11, Springer, 2018.

\bibitem{xiao2015learning}
T.~Xiao, T.~Xia, Y.~Yang, C.~Huang, and X.~Wang,
\newblock Learning from massive noisy labeled data for image classification,
\newblock in {\em Proceedings of the IEEE conference on computer vision and
  pattern recognition}, pages 2691--2699, 2015.

\bibitem{kermany2018identifying}
D.~S. Kermany et~al.,
\newblock Identifying medical diagnoses and treatable diseases by image-based
  deep learning,
\newblock Cell {\bf 172}, 1122--1131 (2018).

\bibitem{li2017webvision}
W.~Li, L.~Wang, W.~Li, E.~Agustsson, and L.~Van~Gool,
\newblock Webvision database: Visual learning and understanding from web data,
\newblock arXiv preprint arXiv:1708.02862  (2017).

\bibitem{lee2018cleannet}
K.-H. Lee, X.~He, L.~Zhang, and L.~Yang,
\newblock Cleannet: Transfer learning for scalable image classifier training
  with label noise,
\newblock in {\em Proceedings of the IEEE Conference on Computer Vision and
  Pattern Recognition}, pages 5447--5456, 2018.

\bibitem{song2019selfie}
H.~Song, M.~Kim, and J.-G. Lee,
\newblock Selfie: Refurbishing unclean samples for robust deep learning,
\newblock in {\em International Conference on Machine Learning}, pages
  5907--5915, PMLR, 2019.

\bibitem{song2020learning}
H.~Song, M.~Kim, D.~Park, Y.~Shin, and J.-G. Lee,
\newblock Learning from noisy labels with deep neural networks: A survey,
\newblock arXiv preprint arXiv:2007.08199  (2020).

\bibitem{park2020provable}
S.~Park, J.~Lee, C.~Yun, and J.~Shin,
\newblock Provable memorization via deep neural networks using sub-linear
  parameters,
\newblock arXiv preprint arXiv:2010.13363  (2020).

\bibitem{jiang2018mentornet}
L.~Jiang, Z.~Zhou, T.~Leung, L.-J. Li, and L.~Fei-Fei,
\newblock Mentornet: Learning data-driven curriculum for very deep neural
  networks on corrupted labels,
\newblock in {\em International Conference on Machine Learning}, pages
  2304--2313, PMLR, 2018.

\bibitem{zhang2016understanding}
C.~Zhang, S.~Bengio, M.~Hardt, B.~Recht, and O.~Vinyals,
\newblock Understanding deep learning requires rethinking generalization,
\newblock arXiv preprint arXiv:1611.03530  (2016).

\bibitem{shorten2019survey}
C.~Shorten and T.~M. Khoshgoftaar,
\newblock A survey on image data augmentation for deep learning,
\newblock Journal of Big Data {\bf 6}, 1--48 (2019).

\bibitem{srivastava2014dropout}
N.~Srivastava, G.~Hinton, A.~Krizhevsky, I.~Sutskever, and R.~Salakhutdinov,
\newblock Dropout: a simple way to prevent neural networks from overfitting,
\newblock The journal of machine learning research {\bf 15}, 1929--1958 (2014).

\bibitem{krogh1992simple}
A.~Krogh and J.~A. Hertz,
\newblock A simple weight decay can improve generalization,
\newblock in {\em Advances in neural information processing systems}, pages
  950--957, 1992.

\bibitem{ioffe2015batch}
S.~Ioffe and C.~Szegedy,
\newblock Batch normalization: Accelerating deep network training by reducing
  internal covariate shift,
\newblock in {\em International conference on machine learning}, pages
  448--456, PMLR, 2015.

\bibitem{yu2019does}
X.~Yu, B.~Han, J.~Yao, G.~Niu, I.~Tsang, and M.~Sugiyama,
\newblock How does disagreement help generalization against label corruption?,
\newblock in {\em International Conference on Machine Learning}, pages
  7164--7173, PMLR, 2019.

\bibitem{nguyen2019self}
D.~T. Nguyen, C.~K. Mummadi, T.~P.~N. Ngo, T.~H.~P. Nguyen, L.~Beggel, and
  T.~Brox,
\newblock Self: Learning to filter noisy labels with self-ensembling,
\newblock arXiv preprint arXiv:1910.01842  (2019).

\bibitem{goldberger2016training}
J.~Goldberger and E.~Ben-Reuven,
\newblock Training deep neural-networks using a noise adaptation layer,
\newblock (2016).

\bibitem{yao2018deep}
J.~Yao, J.~Wang, I.~W. Tsang, Y.~Zhang, J.~Sun, C.~Zhang, and R.~Zhang,
\newblock Deep learning from noisy image labels with quality embedding,
\newblock IEEE Transactions on Image Processing {\bf 28}, 1909--1922 (2018).

\bibitem{ghosh2017robust}
A.~Ghosh, H.~Kumar, and P.~Sastry,
\newblock Robust loss functions under label noise for deep neural networks,
\newblock in {\em Proceedings of the AAAI Conference on Artificial
  Intelligence}, volume~31, 2017.

\bibitem{zhang2018generalized}
Z.~Zhang and M.~R. Sabuncu,
\newblock Generalized cross entropy loss for training deep neural networks with
  noisy labels,
\newblock arXiv preprint arXiv:1805.07836  (2018).

\bibitem{wang2019symmetric}
Y.~Wang, X.~Ma, Z.~Chen, Y.~Luo, J.~Yi, and J.~Bailey,
\newblock Symmetric cross entropy for robust learning with noisy labels,
\newblock in {\em Proceedings of the IEEE/CVF International Conference on
  Computer Vision}, pages 322--330, 2019.

\bibitem{lyu2019curriculum}
Y.~Lyu and I.~W. Tsang,
\newblock Curriculum loss: Robust learning and generalization against label
  corruption,
\newblock arXiv preprint arXiv:1905.10045  (2019).

\bibitem{patrini2017making}
G.~Patrini, A.~Rozza, A.~Krishna~Menon, R.~Nock, and L.~Qu,
\newblock Making deep neural networks robust to label noise: A loss correction
  approach,
\newblock in {\em Proceedings of the IEEE Conference on Computer Vision and
  Pattern Recognition}, pages 1944--1952, 2017.

\bibitem{hendrycks2018using}
D.~Hendrycks, M.~Mazeika, D.~Wilson, and K.~Gimpel,
\newblock Using trusted data to train deep networks on labels corrupted by
  severe noise,
\newblock arXiv preprint arXiv:1802.05300  (2018).

\bibitem{wang2017multiclass}
R.~Wang, T.~Liu, and D.~Tao,
\newblock Multiclass learning with partially corrupted labels,
\newblock IEEE transactions on neural networks and learning systems {\bf 29},
  2568--2580 (2017).

\bibitem{zhang2020learn}
X.~Zhang, K.~Zhou, S.~Wang, F.~Zhang, Z.~Wang, and J.~Liu,
\newblock Learn with noisy data via unsupervised loss correction for weakly
  supervised reading comprehension,
\newblock in {\em Proceedings of the 28th International Conference on
  Computational Linguistics}, pages 2624--2634, 2020.

\bibitem{chang2017active}
H.-S. Chang, E.~Learned-Miller, and A.~McCallum,
\newblock Active bias: Training more accurate neural networks by emphasizing
  high variance samples,
\newblock arXiv preprint arXiv:1704.07433  (2017).

\bibitem{messidor2014messidor}
T.-V. MESSIDOR,
\newblock MESSIDOR: methods to evaluate segmentation and indexing techniques in
  the field of retinal ophthalmology. 2014,
\newblock Available on: http://messidor. crihan. fr/index-en. php Accessed:
  October {\bf 9} (2014).

\bibitem{liu2015classification}
T.~Liu and D.~Tao,
\newblock Classification with noisy labels by importance reweighting,
\newblock IEEE Transactions on pattern analysis and machine intelligence {\bf
  38}, 447--461 (2015).

\bibitem{simonyan2014very}
K.~Simonyan and A.~Zisserman,
\newblock Very deep convolutional networks for large-scale image recognition,
\newblock arXiv preprint arXiv:1409.1556  (2014).

\bibitem{szegedy2016rethinking}
C.~Szegedy, V.~Vanhoucke, S.~Ioffe, J.~Shlens, and Z.~Wojna,
\newblock Rethinking the inception architecture for computer vision,
\newblock in {\em Proceedings of the IEEE conference on computer vision and
  pattern recognition}, pages 2818--2826, 2016.

\bibitem{he2016deep}
K.~He, X.~Zhang, S.~Ren, and J.~Sun,
\newblock Deep residual learning for image recognition,
\newblock in {\em Proceedings of the IEEE conference on computer vision and
  pattern recognition}, pages 770--778, 2016.

\bibitem{frenay2013classification}
B.~Fr{\'e}nay and M.~Verleysen,
\newblock Classification in the presence of label noise: a survey,
\newblock IEEE transactions on neural networks and learning systems {\bf 25},
  845--869 (2013).

\bibitem{kvaalseth1989note}
T.~O. Kv{\aa}lseth,
\newblock Note on Cohen's kappa,
\newblock Psychological reports {\bf 65}, 223--226 (1989).

\bibitem{han2018co}
B.~Han, Q.~Yao, X.~Yu, G.~Niu, M.~Xu, W.~Hu, I.~Tsang, and M.~Sugiyama,
\newblock Co-teaching: Robust training of deep neural networks with extremely
  noisy labels,
\newblock arXiv preprint arXiv:1804.06872  (2018).

\bibitem{wei2020combating}
H.~Wei, L.~Feng, X.~Chen, and B.~An,
\newblock Combating noisy labels by agreement: A joint training method with
  co-regularization,
\newblock in {\em Proceedings of the IEEE/CVF Conference on Computer Vision and
  Pattern Recognition}, pages 13726--13735, 2020.

\bibitem{zheng2020error}
S.~Zheng, P.~Wu, A.~Goswami, M.~Goswami, D.~Metaxas, and C.~Chen,
\newblock Error-bounded correction of noisy labels,
\newblock in {\em International Conference on Machine Learning}, pages
  11447--11457, PMLR, 2020.

\bibitem{xie2020self}
Q.~Xie, M.-T. Luong, E.~Hovy, and Q.~V. Le,
\newblock Self-training with noisy student improves imagenet classification,
\newblock in {\em Proceedings of the IEEE/CVF Conference on Computer Vision and
  Pattern Recognition}, pages 10687--10698, 2020.

\end{thebibliography}



\bibliographystyle{./medphy.bst}    


\section*{Figure Legends as a typed list  }
Fig. 1. Plots depicting performance of optical coherence tomography images classification in the training and validation datasets using tensorboard. The training datasets include corrupted noisy label. Training accuracies are compared for different noisy corruption percentage (0,10\%, 20\%, 30\%, 40\%) (a) with cross-entropy loss plotted against the training epoch (b). Test accuracies are compared (c) with the associated cross-entropy loss (d). Plots are normalized with a smoothing factor 0.6 in order to clearly visualize trends.

Fig. 2. The workflow diagram of DNN with plug-and-play BLRM module. The blue part above indicates the training process of benchmark DNN, and the green part below displays the procedure of refurbishing labels with \LMM~. The training dataset is gradually purified after BLRM working.

Fig. 3. Confusion matrices comparing the training data purity for OCT images classification before (a) and after (b) correction at the noise rate of 40\%.

Fig. 4. The test accuracy with different combinations of window width ($T$) and uncertainty threshold ($\varepsilon$), performed as a grid search to determine the optimized combination of hyperparameters for the OCT dataset (a) and the Messidor dataset (b). 

Fig. 5. Confusion matrices comparing test accuracy for OCT images classification without (a) and with (b) BLRM.  

Fig. 6. The ROC curves using Inception-v3 or Inception v3 integrated with BLRM.  (a) to (d) is at 10\%, 20\%, 30\%, and 40\% noise rate, respectively. Inception v3 with \LMM: red. Inception v3 without \LMM: blue.

Fig. 7. Confusion matrices comparing test accuracy for DR images classification without (a) and with (b) BLRM.

Fig. 8. The best test accuracy of the comparative training methods of anti-noise on medical datasets under different noise settings.

Fig. 9. The best test accuracy of the comparative methods of combating noisy labels on ANIMAL-10N.

Fig. 10. Investigation of self-training with BLRM on the Messidor dataset. The X-axis represents the percentage of training data with ground-truth labels fed to DNNs in the Messidor dataset and Y-axis represents test accuracy of DNNs.

Fig. 11. Examples of class activation maps from Inception-V3 with or without \LMM. (a) and (b) are positive cases where blue bounding boxes mark the lesions on the images, and (c) is a negative case. The second and third columns are CAMs with and without BLRM with predicted labels and probabilities in the upper left corner.

\end{document}